# Uncertainty estimation via ensembles of deep learning models and dropout layers for seismic traces

Giovanni Messuti[1], Ortensia Amoroso[1], Ferdinando Napolitano[1], Mariarosaria Falanga[2], Paolo Capuano[1], Silvia Scarpetta[1]

(1) Department of Physics "E.R. Caianiello", University of Salerno, Fisciano, Italy
(2) Department of Information and Electrical Engineering and Applied Mathematics (DIEM), University of Salerno, Fisciano, Italy

e-mail: g.messuti@studenti.unisa.it, sscarpetta@unisa.it

**Abstract** Deep learning models have demonstrated remarkable success in various fields, including seismology. However, one major challenge in deep learning is the presence of mislabeled examples. Additionally, accurately estimating model uncertainty is another challenge in machine learning. In this study, we develop Convolutional Neural Networks (CNNs) to classify seismic waveforms based on first-motion polarity. We trained multiple CNN models with different settings. We also constructed ensembles of networks to estimate uncertainty. The results showed that each training setting achieved satisfactory performances, with the ensemble method outperforming individual networks in uncertainty estimation. We observe that the uncertainty estimation ability of the ensembles of networks can be enhanced using dropout layers. In addition, comparisons among different training settings revealed that the use of dropout improved the robustness of networks to mislabeled examples.

**Keywords** Uncertainty estimation, Ensemble learning, First-motion polarity, Dropout, Convolutional Neural Networks

## 1.1 Introduction

Deep learning models have achieved remarkable success in various fields, including geophysics and, in particular, in seismology [1]. These models have the ability to learn complex patterns and relationships from large datasets, making them well suited for tasks such as image classification, object detection, and time series forecasting. However, one of the major challenges when dealing with large datasets for deep learning is the occurrence of mislabeled items. A comprehensive and high-quality dataset can be challenging to obtain in many areas of application, including seismology. Mislabeling can impact the model performance, leading to poor generalization results [2]. Detecting and fixing mislabeled items is a challenging task that requires significant effort and expertise. Moreover, it can be difficult to figure out the origin of labeling errors, as they may arise from a variety of sources such as human error or ambiguity in the data.

Another challenge in the field of machine learning is the estimation of model uncertainty [3]. Uncertainty can be categorized into two types: aleatoric and epistemic. Aleatoric uncertainty pertains to the stochastic component of phenomena and cannot be reduced by additional information sources, while epistemic uncertainty refers to a lack of knowledge, for example about the sampling distribution. Calibration methods are important techniques for managing uncertainty. They consist in adjusting the model's predictions to accurately reflect the observed probabilities [4]. Nonetheless, calibration processes leverage solely on known examples. As a result, methods for confidence calibration on the input distribution cannot be used for the detection of out-of-distribution data (i.e., data whose distribution differs from the training data) [5], making these techniques ineffective for estimating epistemic uncertainty.

The concept of *trust* plays a crucial role in this framework. It not only concerns establishing how often a model performs well, but also investigates the examples for which it does [6]. Some studies suggest that the interpretability of models is a prerequisite for trust [7, 8]. Although Deep Neural Networks perform exceptionally well, and calibration methods can enhance the reliability of their predictions with known distribution data, they inherently lack interpretability. Consequently, they struggle to exhibit properties of trustworthiness when confronted with epistemic uncertainty.

In our work, we developed Convolutional Neural Networks (CNNs) to classify seismic waveforms based on first-motion polarity, i.e., the sign of the first oscillation subsequent to the P-wave arrival. First-motion polarities are essential for deriving earthquakes' focal mechanisms [9], which provide valuable information about crustal stress fields and tectonics [10]. We collected our data from the INSTANCE dataset [11], which contains seismic traces labeled with respect to polarity information as 'upward', 'downward', and 'undecidable' (i.e., data where the analyst declared the polarity difficult or impossible to extract). A manual inspection revealed the presence of noisy data, namely waveforms with erroneous labels or unreliable polarity in the dataset. We ran multiple training sessions utilizing various settings while employing the same model architecture. The comparison of these settings showed the improved capability of networks trained with dropouts to disregard noise and focus on the informative features within the dataset.
As a secondary analysis, we constructed an ensemble of networks by incorporating all the previously trained models. We demonstrated the greater capacity of the ensemble method in estimating uncertainty within our undecidable data, as evidenced by comparing the prediction distributions of an individual network to the collective ensemble. Furthermore, we observed a notable enhancement in uncertainty estimation when employing a combination of dropout and ensemble techniques.

### 1.2 Methods

Our datasets are derived from INSTANCE, a large seismic dataset specially compiled for machine learning applications. The dataset contains waveforms labeled with defined polarity (upward or downward) or labeled as undecidable polarity. We trained our network on the vertical component of the waveforms with defined polarity registered by velocimeters, while we used the undecidable polarity waveforms (approximately in number of 923,000) to evaluate uncertainty on out-of-distribution examples from unseen data. To this scope, we divided the approximately 160,000 seismic traces with defined polarity into different subsets, assigning 88% of them to the training set, 6.4% to the validation set and the remaining 5.6% to the test set. A starting visual inspection revealed that approximately 8% of the waveforms represented data with ambiguous polarity information or mislabeled data. To address this issue, we developed a cleaning strategy based on cyclic applications of Self-Organized Maps (SOM) [12], and manual visual inspection [13]. This strategy allowed us to explore the dataset, identifying and selecting mislabeled data or waveforms showing not reliable polarity information. As a result of the cleaning procedure, we created a SOM-cleaned dataset, resulting in the removal of around 10,000 traces from the training, validation, and test sets. An example of data discarded from the original dataset is the group of mislabeled negative polarities shown in Fig. 1. In the following, we study the performance of our CNNs and of Ensembles of CNNs, both using the complete dataset and the SOM-cleaned one.

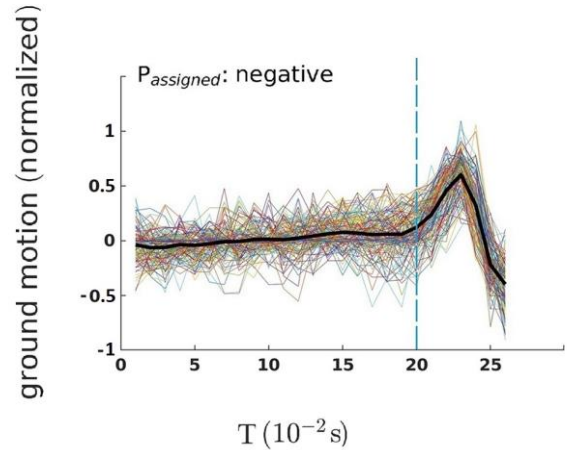

**Figure 1** - Examples of waveforms identified by the SOM visualization technique and excluded by us to build the cleaned dataset. We display the example waveforms (colored lines) and their mean (black line). These waveforms are labeled as 'downward polarity' in the INSTANCE dataset, even if they show a perceptible ascending trend after the arrival. The vertical dashed lines show the declared P-wave arrival samples.

We explored various training settings, testing two different optimizers, and including dropout layers with a drop rate of 0.5. The architecture of each model we trained is represented in Fig. 2. It consists of five convolutional layers and two fully connected layers. Two dropout layers have been applied in only half of the training sessions, while the others remained unchanged among all the training sessions. To carry out our training, we utilized a value of the batch size equal to 512 and the default learning rate value of 0.01 was used (different learning rates in the range [0.007, 0.015] showed similar performances). The momentum parameter for the models trained using the Stochastic Gradient Descent optimizer (SGD) was set to 0.8, while the epsilon parameter when using the ADAM optimizer was set to 0.01. The sigmoid function was employed as the non-linear activation function for the output layer, enabling the interpretation of the output as the probability of the corresponding input waveform to exhibit upward polarity. As a remedy to the overfitting issue, we applied the early stopping technique. We set the maximum epoch to 100 and we interrupted the learning if the loss on the validation set was not improving for 10 consecutive epochs. To increase the amount of training data (and balance the upward/downward labels), we utilized an augmentation strategy, such that, for each seismic trace of the training set, we generated a flipped trace with opposite polarity, by multiplying it by -1.

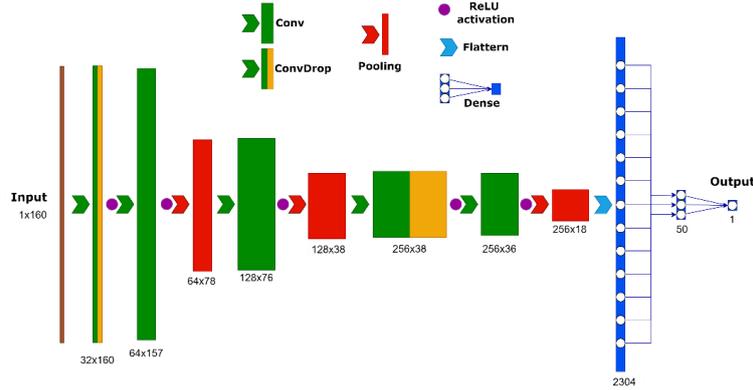

**Figure 2** – Models architecture. Dropout layers are present only in half of training sessions, with a dropout rate of 0.5. Numbers under each layer indicate its shape (number of channels x number of neurons). Maxpooling layer reduces the data dimension by 2. Last layer uses a sigmoid activation function.

### 1.3 Results

We performed eight training sessions by varying the optimizer, the presence of dropout layers and the training dataset (SOM-cleaned or complete dataset). For each training setting, we trained 7 models. The details of the settings related to each training session and their corresponding test performances are provided in Table 1, for the case of the networks trained on the complete dataset, and in Table 2, for the models trained on the SOM-cleaned dataset. Each panel in Table 1 and Table 2 shows the mean accuracy of the 7 independent trainings (with identical settings) on various test sets. We tested the networks both on the complete INSTANCE test set, and on the SOM-cleaned test set, which comprises the same data of the complete test, but where we discarded traces identified by SOM analysis. Test sets are identical for each session, in order to avoid possible fluctuations of performance due to the stochasticity in choosing the test data. We recall that networks are trained using only waveforms with defined polarity.

Results in Table 1 and Table 2 indicate that with the proper hyperparameters choice each training setting can lead to satisfactory performances. Mean accuracies of various training sessions on the same test set show very similar values, in fact, the values are within 2 standard deviations (if we consider the interval estimate of the accuracy using 2 standard deviations as the semi-amplitude of the intervals, we notice that the intersection of all the intervals is not empty). Notably, results indicate that performances on the SOM-cleaned test set are significantly higher for each training. This is a sign that the cleaning procedure has successfully deleted the most ambiguous or mislabeled data points. Despite this, there is no significant difference in performance on the same test set when comparing models trained on the complete training set (Table 1) or the SOM-cleaned training set (Table 2). We confirm that attempting to discard mislabeled examples before the training phase could be ineffective, in some cases, in improving models' performances. This finding is in agreement with studies that have also suggested the possibility of worsening performance if the discarded data exceeds a certain threshold [14].

Training sessions using complete dataset

|  | No Dropout Layers | | Presence of the 2 Dropout Layers | |
|---|---|---|---|---|
| SGD | Instance complete test set | 97.72±0.08 | Instance complete test set | 97.74±0.07 |
|  | Instance SOM-cleaned test set | 98.68±0.07 | Instance SOM-cleaned test set | 98.69±0.04 |
| ADAM | Instance complete test set | 97.76±0.07 | Instance complete test set | 97.77±0.06 |
|  | Instance SOM-cleaned test set | 98.70±0.05 | Instance SOM-cleaned test set | 98.71±0.06 |

**Table 1** – Accuracies of the various training sessions on different test sets training each network on the complete dataset. Right (left) column refers to the architecture presented in Fig. 2 with(out) the presence of the dropout layers. SGD(ADAM) optimizer is used in the training sessions listed in the upper(lower) row.

Training sessions using SOM-cleaned dataset

|  | No Dropout Layers | | Presence of the 2 Dropout Layers | |
|---|---|---|---|---|
| SGD | Instance complete test set | 97.51±0.07 | Instance complete test set | 97.46±0.11 |
|  | Instance SOM-cleaned test set | 98.73±0.04 | Instance SOM-cleaned test set | 98.63±0.09 |
| ADAM | Instance complete test set | 97.61±0.12 | Instance complete test set | 97.51±0.09 |
|  | Instance SOM-cleaned test set | 98.75±0.06 | Instance SOM-cleaned test set | 98,67±0.07 |

**Table 2** – Accuracies of the various training sessions on different test sets training each network on the SOM-cleaned dataset. Right (left) column refers to the architecture presented in Fig. 2 with(out) the presence of the dropout layers. SGD(ADAM) optimizer is used in the training sessions listed in the upper(lower) row.

Despite the similar performances of each training session, further investigation revealed the greater inclination of SGD&dropout training sessions to being robust against overfitting. We formulated this hypothesis after identifying 337 clearly mislabeled data points using the SOM data visualization technique and revising all of them manually. These data showed clear polarity information but the label assigned in the original dataset was wrong (Fig. 3 exhibits a selection of such waveforms). Despite the majority of the mislabeled data belonging to the training set (in the case of the complete dataset), the SGD&dropout training sessions, on average, were able to assign 310 (92%) of them to the correct class. Conversely, other training sessions corrected no more than 290 labels.

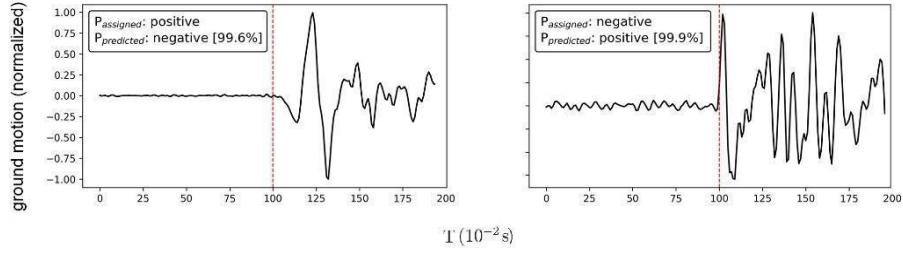

**Figure 3** – Examples of traces assigned with an erroneous label in the INSTANCE dataset. 'P$_{assigned}$' and 'P$_{predicted}$' denote the polarity assigned in the dataset and the predicted polarity, respectively. Predictions shown are performed by one of the networks trained on the complete dataset using SGD&dropout.

### 1.3.1 Ensembles of models: distribution of the predictions

Ensemble methods are widely recognized for their effectiveness in improving the performance of machine learning models [15]. These techniques combine multiple instances of machine learning models to obtain more accurate and generalized predictions. Our study aims to examine how ensemble methods are able to manage uncertainty on seismic traces. Whereas previously we demonstrated the remarkable accuracy of the networks when used to predict the polarity of the test sets, in the following we will analyze the distributions of the networks' predictions, both on traces labeled with a defined polarity, and on the traces where the analyst declared the polarity to be 'undecidable'.

We first compute the mean prediction of all networks on the complete test set with defined polarity (waveforms labeled as upward or downward by an expert analyst in the original INSTANCE dataset). We show the histogram relative to the distribution of the mean predictions in Fig. 4a. At this stage, it is important to remember that the output of the networks can be interpreted as the probability of showing upward polarity. Accordingly, output values close to 1 (0) indicate high confidence of our networks that the input waveform exhibits an upward (downward) polarity. In the same way, we use the undecidable waveforms of the INSTANCE dataset to build the panel in Fig. 4b.

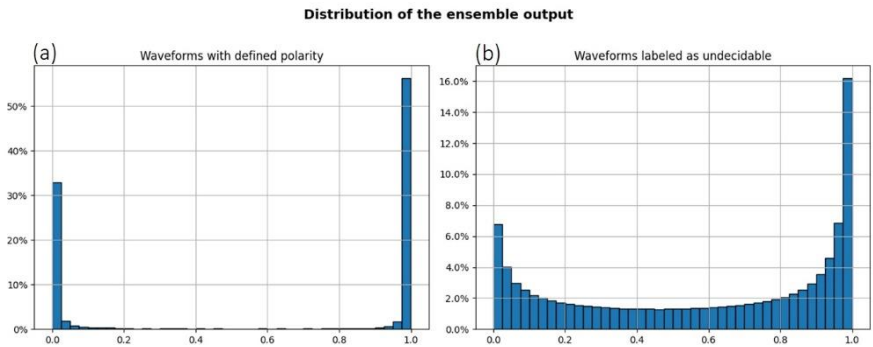

**Figure 4** – Histograms of the mean of predictions of our networks on the complete test set with defined polarity (a) and on undecidable waveforms (b). The mean is computed using each prediction of the 56 networks, whose settings are presented in Table 1 and Table 2. The histogram is divided into 40 bins.

We notice from Fig. 4a that the histogram on defined polarity data shows two peaks, containing more than 90% of data, at the edges of the interval. This is a clear sign of the high confidence of the networks about their predictions. Similar results can be achieved considering the predictions of a single network. Figure 4b shows the ensemble prediction distribution on the undecidable INSTANCE waveforms. Compared to Fig. 4a, the values in the distribution of Fig. 4b appear to be filled more evenly, with the peaks much reduced. This property is desirable, as we expect waveforms exhibiting undecidable polarity to be more likely placed towards the center of the interval [0, 1], i.e., in the region where networks exhibit lower confidence in assigning a polarity.

In Fig. 5, we plot a typical histogram representing the prediction of a single network on undecidable waveforms. By comparing it with the mean distribution shown in Fig. 4b, we observe that the predictions of the single network are more peaked. Although we notice the ensemble placing fewer data in the extremal bins, the distribution on the undecidable waveforms is still peaked on the edges of the interval. We manually analyzed 400 waveforms within the right extremal bin, observing that 140 displayed an upward polarity. Conversely, among the 400 waveforms within the left extremal bin, only 90 displayed a downward polarity, consistent with the ensemble predictions.

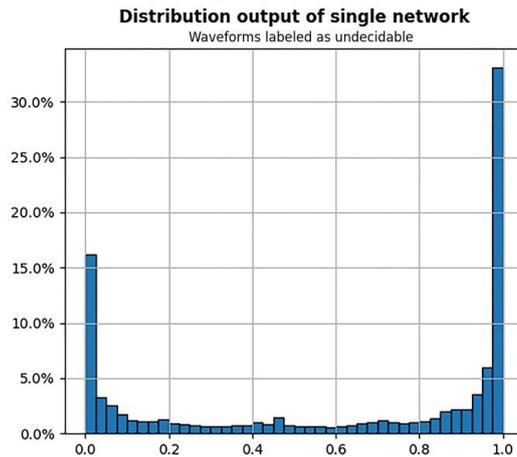

**Figure 5** – Histogram of the distribution of the predictions of a single network on undecidable waveforms.

These observations indicate that while the ensemble outperforms a single network in assessing uncertainty levels, it still demonstrates overconfidence when making predictions on undecidable waveforms, underestimating the epistemic uncertainty deriving from the unknown distribution.

Figures 6 and 7 present in detail the distributions of mean predictions by all the different training sessions conducted, referring to Table 1 and Table 2 respectively.

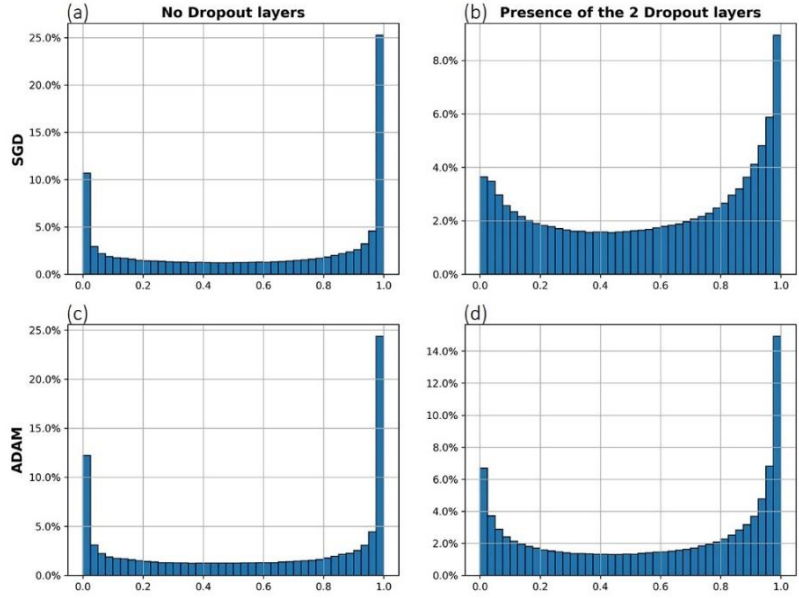

**Figure 6** – Histograms showing the means of predictions on undecidable waveforms, varying different training sessions. Each mean corresponds to a panel in Table 1 (complete INSTANCE training set) and is computed using the respective 7 networks that share the same settings.

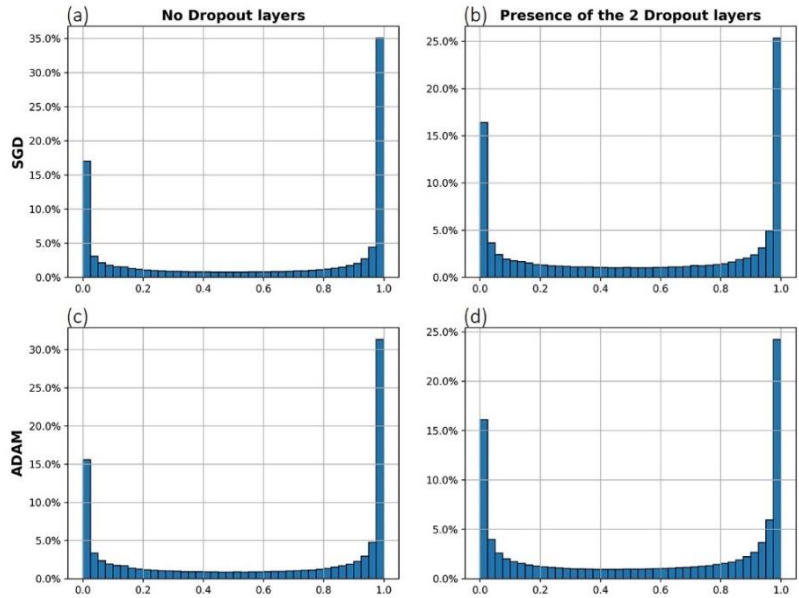

**Figure 7** – Histograms showing the means of predictions on undecidable waveforms, varying different training sessions. Each mean corresponds to a panel in Table 2 (SOM-cleaned training set) and is computed using the respective 7 networks that share the same settings.

We can observe how different settings result in varying degrees of flatness in the distributions. By comparing each panel in Fig. 7 with the respective panel in Fig. 6, we can see that the distributions in Fig. 7 (SOM-cleaned training set) are more peaked. Furthermore, when comparing panels (b) and (d) on the right sides of Fig. 6 and Fig. 7 with their respective panels (a) and (c), on the left of the same figure, we can observe that the use of dropout layers improves the estimation of uncertainty, i. e. placing more data in the central bins of the histograms. The best estimation of uncertainty is achieved when using the complete INSTANCE training set with SGD&dropout setting (Fig. 6b), which even outperforms the ensemble of all the training sessions. It is noteworthy that using dropout layers alone, without employing the ensemble technique, is not sufficient to achieve a reliable estimation of uncertainty. This is because the distribution of predictions from a single network, even with the use of dropout layers, is similar to the one represented in Fig. 5.

### 1.4 Conclusions

Our work involved the training of Convolutional Neural Networks (CNNs) on waveforms labeled with a defined first-motion polarity. Multiple training settings were explored, varying the optimizers used, the inclusion of dropout layers and the training dataset employed. Satisfactory performances in discriminating polarities were achieved across all settings, i.e., accuracies above 97.4%.

A SOM visualization procedure allowed us to identify groups of waveforms with ambiguous and less reliable polarity information. Our results suggest that the cleaning procedure, which involved the exclusion of waveforms identified by the SOM, successfully removed mislabeled and ambiguous data points from the dataset. Despite this, the performances of models trained on the cleaned dataset did not show significantly higher accuracies, indicating the intrinsic robustness of deep learning to noisy data. Notably, we observed the networks trained with dropout layers demonstrate improved capability in handling mislabeled data.

Model uncertainty estimation is another challenge in machine learning. In the framework of single deterministic networks, the distribution over the outputs can be interpreted as a quantification of the model uncertainty. However, these models can often exhibit overconfidence when fed with out-of-distribution examples. Consistently, we observed our networks, trained on defined polarity data, to show overconfident predictions when tested on the unseen undecidable polarity waveforms. To address this issue, we constructed ensembles of CNNs. The ensemble method outperformed individual networks in uncertainty estimation. Interestingly, we discovered that, even if individual models using dropout layers did not show a significant improvement in the distribution of predictions, the uncertainty estimation capability of ensembles incorporating models with dropout layers was significantly enhanced.

In conclusion, in our study:

- We developed various CNNs to determine first-motion polarities. Across various training settings, we achieved accuracies exceeding 97.4%;

- Through a meticulous cleaning procedure, we successfully identified and removed mislabeled and ambiguous data points. Notably, the dropout layers inclusion exhibited effectiveness in addressing the mislabeled data issue;

- We introduced ensemble methods to improve uncertainty estimation. Our findings indicate that ensembles outperformed individual networks, with a significant enhancement observed when incorporating dropout layers.

**Acknowledgements**: This work has been supported by the European Union under the Italian National Recovery and Resilience Plan (NRRP) of NextGenerationEU DM 118/2023.